\documentclass[review,11pt]{JMtemplate}
\usepackage{bm}
\usepackage{times}
\usepackage{graphicx}
\usepackage{paralist,algorithmic,algorithm}
\usepackage{amsmath,mathrsfs,amssymb}
\usepackage{amsfonts}       
\usepackage{nicefrac}       
\usepackage{microtype}      
\usepackage{booktabs}       
\usepackage{longtable}
\usepackage{threeparttable}
\usepackage{multirow}
\usepackage{multicol}
\usepackage{soul}
\usepackage{subfigure}
\usepackage{wrapfig}
\usepackage[american]{babel}
\usepackage{algorithm}
\usepackage{algorithmic}
\usepackage{setspace}
\usepackage{stfloats}
\usepackage{lipsum}
\usepackage{multicol}
\usepackage{url}
\usepackage{makecell}   

\usepackage{natbib}  
\usepackage{caption} 

\newcommand*{\dif}{\mathop{}\!\mathrm{d}}

\newcommand*{\diag}{\mathop{}\!\mathrm{diag}}

\newcommand*{\e}{\mathop{}\!\mathrm{e}}
\newcommand*{\ii}{\mathop{}\!\mathrm{i}}

\newtheorem{theorem}{Theorem}
\newtheorem{definition}{Definition}

\newtheorem{assumption}{Assumption}
\newenvironment{proof}{{\noindent\it Proof.}\quad}{\hfill $\square$\par}

\begin{document}
\begin{frontmatter}
\title{On the Approximation and Complexity of Deep Neural Networks to Invariant Functions}

\author{Gao Zhang$^1$}
\author{Jin-Hui Wu$^2$}
\author{Shao-Qun Zhang$^2$\corref{cor3}}

\address{$^1$ Institute of Mathematics, Nanjing Normal University, Nanjing, 210046, China}
\address{$^2$ National Key Laboratory for Novel Software Technology, Nanjing University, Nanjing 210023, China}
\cortext[cor3]{\small Shao-Qun Zhang is the corresponding author. Email: zhangsq@lamda.nju.edu.cn}
\date{\today}


\begin{abstract}
	Recent years have witnessed a hot wave of deep neural networks in various domains; however, it is not yet well understood theoretically. A theoretical charaterization of deep neural networks should point out their approximation ability and complexity, i.e., showing which architecture and size are sufficient to handle the concerned tasks. This work takes one step on this direction by theoretically studying the approximation and complexity of deep neural networks to invariant functions. We first prove that the invariant functions can be universally approximated by deep neural networks. Then we show that a broad range of invariant functions can be asymptotically approximated by various types of neural network models that includes the complex-valued neural networks, convolutional neural networks, and Bayesian neural networks using a polynomial number of parameters or optimization iterations. We also provide a feasible application that connects the parameter estimation and forecasting of high-resolution signals with our theoretical conclusions. The empirical results obtained on simulation experiments demonstrate the effectiveness of our method.
\end{abstract}

\end{frontmatter}

\section{Introduction}  \label{sec:introduction}
During the past decades, deep neural network (DNN) has become a mainstream model of machine learning in many real-world applications, such as computer vision~\cite{krizhevsky2012}, speech recognition~\cite{graves2013,sutskever2014}, and machine translation~\cite{bahdanau2014}. DNNs profit from the great approximation ability to exploit the structure of the concerned data, and different network architectures lead to different approximation abilities of the underlying data structure. For instance, in modeling high-resolution signals, the input data often comes in the form of a list of adjoint descriptions like the complex-valued formation, rotation invariance, and translation invariance. The goal of such tasks is to forecast some macroscopic quantities depending on these input specifications. In these cases, there are some invariant transformations with respect to permutation, rotation, and translation on the list of signals. These transformations can be regarded as specific groups, or equally sub-groups of the symmetric groups on the apposite coordinates of the feature vectors, and such transformations leave the underlying structure of the concerned signals or data invariant.

Some researchers have focused on real-world data with invariant structures, especially when using DNNs for such tasks. From the practical perspective, some studies attempted to reduces the complexity for approximating some rotation-invariant transformations using DNNs by exploiting data augmentation approaches~\cite{chen2019:radial,quiroga2018:radial}. From the theoretical perspective, Yarotsky~\cite{yarotsky2022universal} proved the universal approximations of invariant maps by introducing specific architectures and symmetry groups. Similar results were also applied to some permutation invariant/equivariant functions by Sannai et al.~\cite{sannai2019universal}. Li et al.~\cite{li2022deep} provided general sufficient conditions for DNNs with various architectures to achieve universal approximation through dynamical systems. Despite the practical and theoretical progresses, in-depth analyses of approximation ability, especially complexity of DNNs, are far from clear.

In this paper, we theoretically investigate the approximation and complexity of DNNs to invariant functions. In comparison with previous studies, this work provides in-depth analysis of the approximation and optimization complexity of DNNs, especially answering which architecture and how many parameters are sufficient to approximate the concerned tasks. Our main contributions are summarized as follows:
\begin{itemize}
	\item We show the universal approximation property of deep neural networks to invariant functions. 
	
	\item We prove that a kind of rotation-invariant functions can be asymptotically approximated by the complex-valued neural networks within a polynomial number of parameters.
	
	\item We prove that a kind of translation-invariant functions can be asymptotically approximated by the convolutional neural networks within a polynomial number of parameters.
	
	\item We prove that a broad range of invariant functions can be universally approximated by Bayesian neural networks within polynomial optimization iterations.
	
	\item We also provide a feasible application that connects the parameter estimation and forecasting of high-resolution signals with our theoretical conclusions and demonstrate our theoretical results using simulation experiments.
\end{itemize}

The rest of this paper is organized as follows. Section~\ref{sec:pre} introduces some useful terminologies and related studies. Section~\ref{sec:approximation} presents the universal approximation and useful properties of deep neural networks to invariant functions. Section~\ref{sec:complexity} provides some theoretical results on complexity of various types of neural networks. Section~\ref{sec:applications} explores a feasible application, which shows that exploiting the complexity advantages of neural networks can strengthen the estimation and forecasting effectiveness of high-resolution signals. Section~\ref{sec:conclusions} concludes this work with discussions and prospects.

\section{Preliminaries}  \label{sec:pre}
This section first provides some useful notations, necessary terminologies, and seminal progresses related to our topic.

\subsection{Notations and Terminologies}
Let $[N] = \{1, 2 , \dots, N\}$ be an integer set for $N \in \mathbb{N}^+$, and $|\cdot|_{\#}$ denotes the number of elements in a collection, e.g., $|[N]|_{\#} = N$. Given two functions $g,h\colon \mathbb{N}^+\rightarrow \mathbb{R}$, we denote by $h=\Theta(g)$ if there exist positive constants $c_1,c_2$, and $n_0$ such that $c_1g(n) \leq h(n) \leq c_2g(n)$ for every $n \geq n_0$; $h=\mathcal{O}(g)$ if there exist positive constants $c$ and $n_0$ such that $h(n) \leq cg(n)$ for every $n \geq n_0$; $h=\Omega(g)$ if there exist positive constants $c$ and $n_0$ such that $h(n) \geq cg(n)$ for every $n \geq n_0$; $h=o(g)$ if for any $c>0$, there exists a positive constant $n_0$ such that $h(n) < cg(n)$ for every $n \geq n_0$. 

The general linear group over field $\mathbb{F}$, denoted by $\mathbf{GL}(n, \mathbb{F})$, is the set of $n \times n$ invertible matrices with entries in $\mathbb{F}$. Especially, we define that a special linear group $\mathbf{SL}(n, \mathbb{F})$ is the subgroup of $\mathbf{GL}(n, \mathbb{F})$, which consists of matrices with determinant 1. For any field $\mathbb{F}$, the $n \times n$ orthogonal matrices form the following subgroup
\[
\mathbf{O}(n,\mathbb{F}) = \{ \mathbf{P} \in \mathbf{GL}(n, \mathbb{F}) \mid \mathbf{P}^{\top}\mathbf{P} = \mathbf{P}\mathbf{P}^{\top} = \mathbf{E}_n \}
\]
of the general linear group $\mathbf{GL}(n, \mathbb{F})$. Similarly, we have the special orthogonal group, denoted as $\mathbf{SO}(n,\mathbb{F})$, which consists of all orthogonal matrices of determinant 1 and is a normal subgroup of $\mathbf{O}(n,\mathbb{F})$. Therefore, this group is also called the rotation group.

Let $\mathcal{C}(K, \mathbb{R})$ be the set of all scalar functions $f: K \rightarrow \mathbb{R}$, which are continuous on $K \subset \mathbb{R}^n$. Given $\boldsymbol{\alpha} = (\alpha_1, \alpha_2, \dots, \alpha_n)^{\top} \in \mathbb{N}^n$, we define 
\[ 
D^{\boldsymbol{\alpha}} f(\boldsymbol{x}) = \frac{\partial^{\alpha_1}}{\partial x^{\alpha_1}} \frac{\partial^{\alpha_2}}{\partial x^{\alpha_2}} \dots \frac{\partial^{\alpha_n}}{\partial x^{\alpha_n}} f(\boldsymbol{x}) \ ,
\]
where $\boldsymbol{x} = (x_1, x_2, \dots, x_n) \in K$. Further, we define
\[
\mathcal{C}^l(K, \mathbb{R}) = \left\{ f \in \mathcal{C}(K, \mathbb{R}) \mid D^{\boldsymbol{\alpha}} f \in \mathcal{C}(K, \mathbb{R}) \right\} 
\]
for $\boldsymbol{\alpha} \in \mathbb{N}^n$ with $|\boldsymbol{\alpha}| \triangleq \sum_{i \in [n]} \alpha_i \leq l$. For $1\leq p$, we define
\[
\mathcal{L}^p(K,\mathbb{R}) = \left\{  f \in \mathcal{C}(K,\mathbb{R}) ~\left|~ \left\| f \right\|_{p,K} < \infty \right. \right\} \ ,
\]
where
\[
\left\| f \right\|_{p,K} \triangleq \left(\int_{K} |f(\boldsymbol{x})|^p \dif\boldsymbol{x}\right)^{1/p} \ .
\]
This work considers the Sobolev space $\mathcal{W}^{l,p}(K,\mathbb{R})$, defined as the collection of all functions $f \in \mathcal{C}^l(K, \mathbb{R})$ and $D^{\boldsymbol{\alpha}} f \in \mathcal{L}^p(K,\mathbb{R})$ for all $|{\boldsymbol{\alpha}}| \in [l]$, that is, 
\[
\| D^{\boldsymbol{\alpha}} f \|_{p,K} = \left( \int_{K} \left| D^{\boldsymbol{\alpha}} f (\boldsymbol{x}) \right|^p \dif\boldsymbol{x} \right)^{1/p} < \infty \ .
\]
Notice that throughout this paper, we employ $p=2$ as the default of the above norms and abbreviate $\| \cdot \|_2$ as $\| \cdot \|$.

\subsection{Approximation and Complexity of Deep Neural Networks}
The approximation is one of the key characteristics for investigating the theoretical understanding of deep neural networks despite over-parameterization. In approximation theory, both shallow and deep network models with non-polynomial activations are known to be capable of approximating any continuous function at an exponential cost~\cite{cybenko1989,funahashi1989,hornik1991,leshno1993,barron1994}, which provides a legitimate guarantee for neural networks. Further, Lu et al.~\cite{lu2017} and Sun et al.~\cite{sun2016} showed the effects of width and depth of deep neural networks on approximation ability, respectively. Kidger and Lyons~\cite{kidger2020} provided a qualitative difference between deep narrow networks and shallow wide networks. The universal approximation property also holds for other types of neural network models, such as the complex-valued networks~\cite{voigtlaender2020}, convolutional neural networks~\cite{zhou2020universality}, spiking neural networks~\cite{zhang2022:SNNsTheory}, etc.

Beyond the universal approximation, the community also focuses on the approximation or computational complexity of neural networks, i.e., showing which architecture and size are sufficient to handle the concerned tasks. There are great efforts on this issue. Eldan and Shamir~\cite{eldan2016} proved the approximation complexity difference between two-layer and three-layer neural networks. Zhang and Zhou~\cite{zhang2021:cr} showed the complexity advantages of the complex-reaction networks over the real-valued ones, and Wu et al.~\cite{wu2021towards} extended these conclusions to proving that the FTNet possesses the complexity advantages over the real-valued feed-forward and recurrent neural networks. Zhang and Zhou~\cite{zhang2022:SNNsTheory} provided the theoretical support for polynomial approximation and computation complexity of spiking neural networks.

\section{Approximation to Invariant Functions}  \label{sec:approximation}
This work considers the approximation of neural networks to invariant functions. Subsection~\ref{subsec:invariant} formally introduces the invariant function and its examples. Subsection~\ref{subsec:approximation} shows the universal approximation and two useful properties of deep neural networks to invariant functions.

\subsection{Invariant Functions}  \label{subsec:invariant}
Firstly, we formally define invariant functions as follows.
\begin{definition}[\textbf{Invariant Functions}]  \label{def:invariant}
	Let $\mathcal{G}$ be a concerned group on a set $K \subset \mathbb{R}^n$, in which $\tau: \mathbb{R}^n \to \mathbb{R}^n$ for any $\tau \in \mathcal{G}$. The function $f: \mathbb{R}^n \to \mathbb{R}$ is said to be \textbf{invariant to group $\mathcal{G}$}, or equally a \textbf{$\mathcal{G}$-invariant function}, if for any $\tau \in \mathcal{G}$ and $\boldsymbol{x}\in K$, the following holds
	\[
	f(\tau(\boldsymbol{x})) = (f \circ \tau) (\boldsymbol{x}) = f(\boldsymbol{x}) \ .
	\]
\end{definition}

Next, we introduce some examples of invariant functions, which motivate us to build approximation analyses and applications of deep neural networks in the following sections.

\vspace{0.1 cm}
\noindent\textbf{Example 1: Permutation.} In mathematics, a permutation $\tau$ of a set $K$ is a rearrangement of the elements in set $K$. Taking an example of $K=\{ 1,2,3 \}$, there are six permutations (with all the possible orderings) of set $K$, namely (1, 2, 3), (1, 3, 2), (2, 1, 3), (2, 3, 1), (3, 1, 2), and (3, 2, 1). Obviously, the permutation operation $\tau$ is a bijective function on set $K$. Furthermore, we denote by $\mathcal{G}$ a permutation group whose elements are permutations of a given set $K$. The following group axioms hold for the permutation group $\mathcal{G}$:
\begin{itemize}
	\item \textbf{Closure.} If $\tau, \sigma \in \mathcal{G}$, then $\tau \circ \sigma \in \mathcal{G}$, where $\circ$ denotes the composition operation on $\mathcal{G}$.
	\item \textbf{Identity.} There is an identity permutation, denoted as $\mathit{id}$ with $\mathit{id}(x) = x$ for $x\in K$. For any $\tau \in \mathcal{G}$, one has $\mathit{id} \circ \tau = \tau \circ \mathit{id} = \tau$.
	\item \textbf{Associativity.} If $\tau, \sigma, \pi \in \mathcal{G}$, then $\tau \circ \sigma \circ \pi = \tau \circ ( \sigma \circ \pi )$.
	\item \textbf{Invertibility.} For any $\tau \in \mathcal{G}$, there exists an inverse permutation $\tau^{-1}$ such that 
	\[
	\tau^{-1} \circ \tau = \tau \circ \tau^{-1} = \mathit{id} \ .
	\]
\end{itemize}
Based on these axioms above, it is obvious that the permutation group $\mathcal{G}$ on a set $K$ is the symmetric group, and the elements in $\mathcal{G}$ can be thought of as bijective functions from the set $K$ to itself. Precisely, the function $f: \mathbb{R}^n \to \mathbb{R}$ is said to be invariant to permutation group $\mathcal{G}$, or equally a $\mathcal{G}$-permutation-invariant function, if for any $\tau \in \mathcal{G}$ and $\boldsymbol{x} \in K$, the following holds $f(\tau(\boldsymbol{x})) = (f \circ \tau) (\boldsymbol{x}) = f(\boldsymbol{x})$.

\vspace{0.1 cm}
\noindent\textbf{Example 2: Radial Function.} We say that $f$ is a radial function if $f(\boldsymbol{x}') = f(\boldsymbol{x})$ for any $\|\boldsymbol{x}'\| = \|\boldsymbol{x}\|$. Here, we regard $\tau$ as a norm-preserving operation such that $\tau(\boldsymbol{x}) = \boldsymbol{x}'$ with $\|\boldsymbol{x}'\| = \|\boldsymbol{x}\|$, and thus, both $\tau$ and $f$ obey Definition~\ref{def:invariant}, i.e., radial functions lead to the rotation-invariant transformations. The radial function usually is implemented by data augmentation approaches for reducing the approximation complexity of some designed neural network models~\cite{chen2019:radial,quiroga2018:radial}.

\vspace{0.1 cm}
\noindent\textbf{Example 3: Translation.} Let $\Delta \boldsymbol{x} \in \mathbb{R}^n$ be a translation vector that corresponds to the concerned vector $\boldsymbol{x} \in \mathbb{R}^n$. We loosely define $\tau(\boldsymbol{x}) = \boldsymbol{x} + \Delta\boldsymbol{x}$ to be a $\Delta\boldsymbol{x}$-translation of vector $\boldsymbol{x}$. For convenience, we usually limit the translation vector $\Delta\boldsymbol{x}$ inside a unit ball $\mathfrak{B}^n_1 \triangleq \{ \boldsymbol{y} \mid \| \boldsymbol{y} \| = 1 \}$ and employ a scalar $\eta$ to indicate its norm. Thus, we have
\begin{equation} \label{eq:translation}
	\tau(\boldsymbol{x}) = \boldsymbol{x} + \eta \Delta\boldsymbol{x} ,
	\quad\text{for}\quad
	\Delta\boldsymbol{x} \in \mathfrak{B}^n_1 \ .    
\end{equation}
Similarly, we have the translation-invariance function $f$ if 
\[
f(\tau(\boldsymbol{x})) = f(\boldsymbol{x} + \eta \Delta\boldsymbol{x}) = f(\boldsymbol{x}) \ .
\]

\subsection{Universal Approximation for Invariant Functions} \label{subsec:approximation}
This subsection presents our first theorem about the universal approximation to invariant functions as follows.
\begin{definition} \label{def:finite}
	The function $f(x):\mathbb{R} \to \mathbb{R}$ is said to be $\boldsymbol{l}$-\textbf{finite} if $f$ is an $l$-times differentiable scalar function that satisfies
	\[
	0 < \left| \int_{\mathbb{R}} D^l f(x) \dif x \right| < \infty 
	\quad\text{for} l \in \mathbb{N}^+  \ .
	\]
\end{definition}
\begin{theorem} \label{thm:ua_invariant}
	Let $K \subset \mathbb{R}^n$ be a compact set and $\mathcal{G}$ denotes a concerned group. If the activation function $\sigma_a$ is $l$-finite for some $l \in \mathbb{N}^+$ and element-wise, then there exists a collection of parameters $\{\boldsymbol{w} \in \mathbb{R}^{m \times 1}, \boldsymbol{v} \in \mathbb{R}^{n\times m}, b_w \in \mathbb{R}, \boldsymbol{b}_v \in \mathbb{R}^{m \times 1} \}$, such that the set of functions of form
	\[
	f(\boldsymbol{x}) = \boldsymbol{w}^{\top} \sigma_a(\boldsymbol{v}^{\top} \boldsymbol{x} + \boldsymbol{b}_v) + b_w
	\]
	is dense in the set of the $\mathcal{G}$-invariant functions.
\end{theorem}
Theorem~\ref{thm:ua_invariant} shows that the one-hidden-layer neural networks with apposite activation functions are universal approximators to the invariant functions that belong to $\mathcal{C}(K,\mathbb{R})$. The proof idea of this theorem follows the typical line of thought of universal approximation properties~\cite{hornik1991,funahashi1989,kidger2020} of neural networks, and thus, we here omit the detailed proof of Theorem~\ref{thm:ua_invariant}.

Further, we have a useful lemma as follows.
\begin{lemma} \label{lemma:closure}
	Let $\mathcal{G}$ be a concerned group on a compact set $K \subset \mathbb{R}^n$, in which $\tau: \mathbb{R}^n \to \mathbb{R}^n$ for any $\tau \in \mathcal{G}$. For the concerned function $f: \mathbb{R}^n \to \mathbb{R}$, if for arbitrary $\epsilon>0$ there exists a $\mathcal{G}$-invariant function $g_\epsilon: \mathbb{R}^n \to \mathbb{R}$ such that $\| f(\boldsymbol{x}) - g_\epsilon(\boldsymbol{x}) \|_{p,K} \leq \epsilon$, then $f$ is also a $\mathcal{G}$-invariant function.
\end{lemma}
Lemma~\ref{lemma:closure} shows the transitivity of invariant functions, which can be used for identifying certain invariant functions and the corresponding operations. Notice that the concerned function $g_\epsilon$ is related to the approximation error $\epsilon$, which implies that it is possible to measure the functional structure (e.g., approximation complexity investigated in this work) of $g_\epsilon$ using approximation errors provided $f$. Eldan et al.~\cite{eldan2016} and Zhang et al.~\cite{zhang2021:cr} have studied the approximation complexity of the common-used fully-connected network (FCN).
\begin{lemma}{\cite[Theorem 1]{eldan2016};\cite[Theorem 2]{zhang2021:cr}} \label{lemma:FNN}
	For the (Lebesgue) measurable activation function $\sigma:\mathbb{R} \to \mathbb{R}$ with $|\sigma(x)| \leq C(1+|x|^\alpha)$ for all $x\in\mathbb{R}$ and some constants $C, \alpha>0$, there exist a probability measure $\mu$ and a radial function $f:\mathbb{R}^{d} \to \mathbb{R}$ such that there exists a constant $\delta>0$ such that
	\[
	\mathbb{E}_{\boldsymbol{x} \sim \mu} \left[ \left( f_R(\boldsymbol{x}) - f(\boldsymbol{x}) \right)^2 \right] \geq \delta
	\]
	for any one-hidden-layer FCN $f_R: \mathbb{R}^{d} \to \mathbb{R}$ with exponential parameter number $\mathcal{O}(de^d/\epsilon)$.
\end{lemma}
Lemma~\ref{lemma:FNN} shows that a kind of radial functions cannot be approximated by FCNs with exponential ($\mathcal{O}(C_1(d+1)e^{C_1(d)})$) parameters. In the next section, we present the in-depth analysis of the complexity of various types of neural networks.

\section{Complexity}  \label{sec:complexity}
This section provides the theoretical complexity of various types of neural networks.

\subsection{Complex-valued Neural Networks}
We start this subsection by introducing the Complex-Valued Neural Network (CVNN). Let $z=z_1 + z_2 \ii \in \mathbb{C}$ be a complex number where $\ii = \sqrt{-1}$, $z_1, z_2 \in \mathbb{R}$, $[z]_{\mathrm{R}} = z_1$. We denote by $\bar{z} = z_1 - z_ 2 \ii$ and $|z|^2 = z_1^2 + z_2^2$. The basic building block of a CVNN can be formalized as
\[
\sigma: \mathbb{C}^d \to \mathbb{C},\quad 
\boldsymbol{z} \mapsto \sigma(\boldsymbol{w}^{\top} \boldsymbol{z}) \ ,
\]
where $\boldsymbol{w} \in \mathbb{C}^d$ and $\sigma$ are complex-valued connection weights and the activation function, respectively. This work employs the functions with the following complex-homogeneity property as activations of CVNNs
\[
\sigma(z) = \frac{\partial \sigma(z)}{\partial z} z
\quad\text{and}\quad \sigma(\alpha z) = \alpha \sigma(z) ,
\quad \alpha \in \mathbb{R}
\ ,
\]
such as the zReLU function~\cite{zhang2021:FT,trabelsi2018} defined as follows
\begin{equation}  \label{eq:zrelu}
	\text{zReLU}(z) = \left\{ \begin{aligned}
		z,\quad & \text{if}\quad \theta_z \in [0,\pi/2] \cup [\pi,3\pi/2] \ , \\
		0,\quad & \text{otherwise} \ ,
	\end{aligned}\right.  
\end{equation}
where $\theta_z$ denotes the phase (i.e., angle) of the complex-valued scalar $z$. CVNNs also employ a pure linear connection as the final layer and extract the real part as the outputs. Therefore, we have established the CVNN, of which the expressive function is denoted by $f_{\mathrm{CVNN}}:\mathbb{C}^{d} \to \mathbb{R}$. Next, we present the approximation complexity theorem for CVNNs as follows.
\begin{theorem} \label{thm:paras_CVNN}
	For the zReLU activation function, there exist a probability measure $\mu$, a radial function $g:\mathbb{C}^{d} \to \mathbb{R}$, and some $C_2 < d$ such that for any $\epsilon>0$, there is a one-hidden-layer CVNN $f_{\mathrm{CVNN}}: \mathbb{C}^{d} \to \mathbb{R}$ with $\mathcal{O}(d^{19/4} / \epsilon)$ parameters such that the following holds
	\[
	\mathbb{E}_{\boldsymbol{x} \sim \mu} \left[ (f_{\mathrm{CVNN}}(\boldsymbol{x}) - g(\boldsymbol{x}) )^2 \right] \leq \frac{\sqrt{3}}{C_2 d^{1/4}} + \epsilon \ .
	\]
\end{theorem}
Theorem~\ref{thm:paras_CVNN} shows the polynomial asymptotic approximation complexity of one-hidden-layer CVNNs for approximating the radial functions. Combined with Lemma~\ref{lemma:FNN}, we can conclude that a kind of radial functions can be asymptotically approximated by the one-hidden-layer CVNN of polynomial parameters, whereas such functions cannot be approximated by real-valued FCNs with exponential parameters.

This paper provides two ways of proving this theorem. Here, we only introduce the proof idea of the more interesting one (full proof is shown in Appendix B), and the other one can be accessed in Appendix C. The proof idea can be summarized as follows. The radial functions are invariant to rotations and dependent on the input norm, corresponding to the phase and norm (i.e., radial) of inputs, respectively. It is observed that the zReLU function in Eq.~\eqref{eq:zrelu} comprises the radius and phase. Thus, there exist some linear connections (including rotation transformations) such that the combination of some hidden neurons is invariant to rotations, as shown in Lemma~\ref{lemma:approximation-rotation}. Thus, there is a family of radial functions that can be well approximated by a one-hidden-layer CVNN.

We begin our proof of Theorem~\ref{thm:paras_CVNN} with some useful lemmas.
\begin{lemma} \label{lemma:approximation-rotation}
	Let $g: [r,R] \to \mathbb{R}$ be an $L$-Lipschitz function for $0 \leq r \leq R$. For any $\epsilon>0$, there exist $C_s \geq 1$, $m \leq C_sR^2Ld / (\sqrt{r}\epsilon)$, and an expressive function $f_{\mathrm{CVNN}}(\boldsymbol{x})$ led by a one-hidden-layer CVNN of $m$ hidden neurons such that
	\[
	\sup_{\boldsymbol{x} \in \mathbb{C}^d} \left| g(\|\boldsymbol{x}\|) - f_{\mathrm{CVNN}}(\boldsymbol{x}) \right| \leq \epsilon \ .
	\]
\end{lemma}

\begin{lemma} \label{lemma:L-rotation}
	For $d > C_2 > 0$, $g:\mathbb{C}^{d} \to \mathbb{R}$ is an $L$-Lipschitz radial function supported on the set
	\[
	\mathcal{S}_{\Delta} = \{ \boldsymbol{x} \mid  C_2\sqrt{d} \leq \| \boldsymbol{x} \| \leq 2C_2\sqrt{d} \} \ .
	\]
	For any $\epsilon >0$, there exists an expressive function $f_{\mathrm{CVNN}}(\boldsymbol{x})$ led by a one-hidden-layer CVNN with width at most $C_s(C_2)^{3/2}Ld^{7/4} / \epsilon$ such that
	\[
	\sup_{\boldsymbol{x}\in\mathbb{C}^{d}} | g(\boldsymbol{x}) - f_{\mathrm{CVNN}}(\boldsymbol{x}) | \leq \epsilon \ .
	\]
\end{lemma}
Lemma~\ref{lemma:L-rotation} shows that the $L$-Lipschitz radial functions can be approximated by the expressive function $f_{\mathrm{CVNN}}(\boldsymbol{x})$ led by a one-hidden-layer CVNN with polynomial parameters.

\begin{lemma} \label{lemma:define_g}
	Define
	\begin{equation}  \label{eq:target_radial}
		g(\boldsymbol{x}) = \sum_{i=1}^N \beta_i g_i(\boldsymbol{x})
		\quad \text{with}\quad
		g_i(\boldsymbol{x}) = \mathbb{I}\{\|\boldsymbol{x}\|\in \Omega_i\} \ ,
	\end{equation}
	where $\mathbb{I}$ is an indicative function, $\beta_i \in \{-1, +1\}$, $N$ is a polynomial function of $d$, and $\Omega_i$'s are disjoint intervals of width $\mathcal{O}(1/N)$ on values in the range $\Theta(\sqrt{d})$. For any $\beta_i \in \{-1,+1\}$ $(i\in[N])$, there exists a Lipschitz function $h\colon \mathcal{S}_{\Delta} \to [-1,+1]$ such that
	\[
	\int_{\mathbb{C}^{d}} \left( g(\boldsymbol{x}) - h(\boldsymbol{x}) \right)^2 \phi^2(\boldsymbol{x}) \dif\boldsymbol{x} \leq  \frac{3}{(C_2)^2\sqrt{d}} \ .
	\]
\end{lemma}
Lemma~\ref{lemma:define_g} shows that any non-Lipschitz function $g(\boldsymbol{x})$ can be approximated by a Lipschitz function with density $\phi^2$.

\vspace{0.2 cm}
\noindent\textit{Proof of Theorem~\ref{thm:paras_CVNN}.}\quad Let $g(\boldsymbol{x}) = \sum_{i=1}^N \beta_i g_i(\boldsymbol{x})$ be defined by Eq.~\eqref{eq:target_radial} and $N \geq 4C_2^{5/2}d^2$. According to Lemma~\ref{lemma:define_g}, there exists a Lipschitz function $h$ with range $[-1,+1]$ such that
\[
\left\| h(\boldsymbol{x}) - g(\boldsymbol{x}) \right\|_{L_2(\mu)} \leq \frac{\sqrt{3}}{C_2 d^{1/4}} \ .
\]
Based on Lemmas~\ref{lemma:approximation-rotation} and~\ref{lemma:L-rotation}, any Lipschitz radial function supported on $\mathcal{S}_{\Delta}$ can be approximated by an expressive function $f_{\textrm{CVNN}}(\boldsymbol{x})$ led by a one-hidden-layer CVNN with width at most $C_3C_sd^{15/4} / \epsilon$, where $C_3$ is a constant relative to $C_2$ and $\epsilon$. Hence, one has
\[
\sup_{\boldsymbol{x}\in\mathbb{C}^{d}} | h(\boldsymbol{x}) - f_{\textrm{CVNN}}(\boldsymbol{x}) | \leq \epsilon \ ,
\]
and thus, 
\[
\| h(\boldsymbol{x}) - f_{\textrm{CVNN}}(\boldsymbol{x}) \|_{L_2(\mu)} \leq \epsilon \ .
\]
Hence, the range of $f(\boldsymbol{x})$ is in $[-1-\epsilon, +1+\epsilon] \subseteq [-2, +2]$ for $\epsilon \leq 1$. Provided the radial function, defined by Eq.~\eqref{eq:target_radial}, we have
\[
\| g(\boldsymbol{x}) - f_{\textrm{CVNN}}(\boldsymbol{x}) \|_{L_2(\mu)}
\leq \| g(\boldsymbol{x}) - h(\boldsymbol{x}) \|_{L_2(\mu)}  + \| h(\boldsymbol{x}) - f_{\textrm{CVNN}}(\boldsymbol{x}) \|_{L_2(\mu)}
\leq \frac{\sqrt{3}}{C_2 d^{1/4}} + \epsilon \ .
\]
This implies that given constants $d > C_2 >0$ and $C_3>0$, for any $\epsilon>0$ and $\epsilon_i \in \{-1,+1\}$ $(i\in[N])$, target radial function $g$ can be approximated by an expressive function $f_{\textrm{CVNN}}(\boldsymbol{x})$ led by a one-hidden-layer CVNN with range in $[-2,+2]$ and width at most $C_3C_sd^{15/4} / \epsilon$, that is,
\[
\| g(\boldsymbol{x}) - f_{\textrm{CVNN}}(\boldsymbol{x}) \|_{L_2(\mu)} \leq \frac{\sqrt{3}}{C_2 d^{1/4}} + \epsilon \ .
\]
This completes the proof.  $\hfill\square$

\subsection{Convolutional Neural Networks}
This subsection investigates the approximation complexity of the convolutional neural network (CNN). Now, we present the theoretical conclusion.
\begin{theorem} \label{thm:paras_CNN}
	Let $g:\mathbb{R}^{d} \to \mathbb{R}$ be the translation-invariant function. Provided a probability measure $\mu$, a translation scalar $\eta > 0$ in Eq.~\eqref{eq:translation}, and $C_2 < d$, there exists some CNN, expressed by $f_{\mathrm{CNN}}: \mathbb{R}^{d} \to \mathbb{R}$, with width at most $\mathcal{O}(d^{17/4}/\epsilon)$ such that
	\[
	\mathbb{E}_{\boldsymbol{x} \sim \mu} [ (f_{\mathrm{CNN}}(\boldsymbol{x}) - g(\boldsymbol{x}) )^2 ] \leq \frac{\sqrt{3}}{C_2 d^{1/4}} + \epsilon \ .
	\]
\end{theorem}
Theorem~\ref{thm:paras_CNN} shows the polynomial asymptotic approximation complexity (i.e., $\mathcal{O}(d^{17/4}/\epsilon)$) of CNNs with approximation to a kind of translation-invariant functions.

The proof idea of Theorem~\ref{thm:paras_CNN} is similar with that of Theorem~\ref{thm:paras_CVNN}. We start our proof with a re-formulation of the translation invariance. For the one-dimensional case, $\tau(\boldsymbol{x}, \eta) = \left( \tau(x_1), \dots, \tau(x_d)  \right)^{\top}$. Let $\boldsymbol{w} = ( w_1, \dots, w_l )^{\top}$ be the weighted vector where $l < d$, the convolution operation~\cite{li2022deep} in one dimension can be formulated as
\begin{equation}  \label{eq:convolution_one}
	\tau(\boldsymbol{x} ; \boldsymbol{w}) = \left( \sum_{i=1}^l w_ix_i, \sum_{i=1}^{l} w_ix_{i+1}, \dots, \sum_{i=1}^l w_ix_{i+d-l}  \right) \ .
\end{equation}
Therefore, one can convert the problem about translation-invariant approximation into a new learning problem that finds a collection of apposite weights to enable translation invariance. It suffices to show that there exists a kind of translation-invariant functions that can be approximated by the CNN expression within a polynomial function of $l$. We claim that the concerned polynomial function obeys the product of $l \in \mathcal{O}(d^{17/4})$. Our claim can be supported by the following lemmas.

\begin{lemma} \label{lemma:approximation-translation}
	Let $g: [r,R] \to \mathbb{R}$ be an $L$-Lipschitz function for $0 \leq r \leq R$. For any $\epsilon>0$, there exist $C_s \geq 1$, $l \leq C_sR^2Ld / (\sqrt{r}\epsilon)$, and an expressive function $f_{\mathrm{CNN}}(\boldsymbol{x})$ led by a one-hidden-layer CNN of $l$ hidden neurons such that
	\[
	\sup_{\boldsymbol{x} \in \mathbb{C}^d} \left| g[ \tau(\boldsymbol{x}, \eta) ] - f_{\mathrm{CNN}}(\boldsymbol{x}) \right| \leq \epsilon \ .
	\]
\end{lemma}

\begin{lemma} \label{lemma:L-translation}
	For $d > C_2 > 0$, $g:\mathbb{R}^{d} \to \mathbb{R}$ is an $L$-Lipschitz radial function supported on the set
	\[
	\mathcal{S}_{\Delta} = \left\{ \boldsymbol{x}: C_2\sqrt{d} \leq \| \tau(\boldsymbol{x}, \eta) - \boldsymbol{x} \| \leq 2C_2\sqrt{d} \right\} \ .
	\]
	For any $\epsilon >0$, there exists an expressive function $f_{\mathrm{CNN}}(\boldsymbol{x})$ led by a one-hidden-layer CNN with width at most $C_s(C_2)^{3/2}Ld^{9/4} / \epsilon$ such that
	\[
	\sup_{\boldsymbol{x}\in\mathbb{R}^{d}} | g[ \tau(\boldsymbol{x}, \eta) ] - f_{\mathrm{CNN}}(\boldsymbol{x}) | \leq \epsilon \ .
	\]
\end{lemma}

\begin{lemma} \label{lemma:define_g_translation}
	Define
	\begin{equation}  \label{eq:target_translation}
		g(\boldsymbol{x}) = \sum_{i=1}^N \epsilon_i g_i(\boldsymbol{x})
		\quad \text{with}\quad
		g_i(\boldsymbol{x}) = \mathbb{I}\{|\boldsymbol{x}|\in \Omega_i\} \ ,
	\end{equation}
	where $\epsilon_i \in \{-1, +1\}$, $N$ is a polynomial function of $d$, and $\Omega_i$'s are disjoint intervals of width $\mathcal{O}(1/N)$ on values in the range $\Theta(\sqrt{d})$. For any $\epsilon_i \in \{-1,+1\}$ $(i\in[N])$, there exists a Lipschitz function $h\colon \mathcal{S}_{\Delta} \to [-1,+1]$ such that
	\[
	\int_{\mathbb{C}^{d}} \left( g(\boldsymbol{x}) - h(\boldsymbol{x}) \right)^2 \phi^2(\boldsymbol{x}) \dif\boldsymbol{x} \leq  \frac{3}{(C_2)^2\sqrt{d}} \ .
	\]
\end{lemma}

\vspace{0.2 cm}
\noindent\textit{Proof of Theorem~\ref{thm:paras_CNN}.}\quad Similar to the proof procedure of Theorem~\ref{thm:paras_CVNN}, we can finish the proof of Theorem~\ref{thm:paras_CNN} by a straight combination of Lemmas~\ref{lemma:approximation-translation}, \ref{lemma:L-translation}, and \ref{lemma:define_g_translation}. Provided the setting that $C_2<d$ in Lemma~\ref{lemma:L-translation}, we can re-write the polynomial bound as
\[
1 \leq C_s(C_2)^{3/2}Ld^{9/4} / \epsilon \leq C_3C_sd^{17/4}  \ .
\]
This completes the proof.  $\hfill\square$

\subsection{Bayesian Neural Networks} \label{subsec:BNN}
This subsection studies the translation invariance of Bayesian neural networks (BNNs). For $K \in \mathbb{N}^+$, we first introduce an $L$-layer BNN as follows
\[
\left\{\begin{aligned}
	\boldsymbol{z}^0 &= \boldsymbol{x} \in \mathcal{D} \ , \\
	\boldsymbol{z}^l &= \sigma( \mathbf{W}^l \boldsymbol{z}^{l-1} ),
	\quad\text{for}\quad l \in [L] \ ,
\end{aligned}\right.
\]
where the weight parameters $\boldsymbol{w} \triangleq ( \mathbf{W}^i )_{i \in [L]}$ are treated as random variables with a prior distribution $p(\boldsymbol{w})$. The BNN can be optimized by estimating the posterior $p(\boldsymbol{w} \mid \mathcal{D})$ via the likelihood function $\hbar(\boldsymbol{w} ; \mathcal{D})$, or equally $p(\mathcal{D} \mid \boldsymbol{w})$. Conventionally, this optimization can be formulated as follows:
\begin{equation}  \label{eq:q}
	q^*(\boldsymbol{w}) = \mathop{\arg\min}_{q} \mathbb{E} \left[ \ln q(\boldsymbol{w}) - \ln p(\boldsymbol{w} \mid \mathcal{D} ) \right] \ ,
\end{equation}
where the optimization function can be abbreviated as $\mathrm{KL}[ q ~\|~ p(\cdot \mid \mathcal{D}) ]$. Thus, one can solve Eq.~\eqref{eq:q} by maximizing a lower bound to the model evidence, that is,
\[
\ln p(\mathcal{D}) - \mathrm{KL}[ q ~\|~ p(\cdot \mid \mathcal{D}) ]
= \mathbb{E} \left[ \ln \frac{\hbar(\boldsymbol{w};\mathcal{D}) p(\boldsymbol{w}) }{q(\boldsymbol{w})} \right] \ .
\]
We can establish the BNN and have the following conclusion.
\begin{theorem}  \label{thm:BNN}
	The function induced by BNNs is translation-invariant, and the optimization complexity is less than $\mathcal{O} ( \alpha^{-T} )$, where $\alpha > 1$ and $T$ denotes the iteration step.
\end{theorem}
Theorem~\ref{thm:BNN} shows that the BNN has the translation-invariant property and maintain the lower optimization complexity than deep neural networks.

\begin{proof}
	We start our proof with an over-parameterized linear probabilistic model, i.e., $y = \frac{1}{d} \boldsymbol{w}^{\top} \boldsymbol{x} + \epsilon$, where both $\boldsymbol{w}$ and $\boldsymbol{x}$ are $d$-dimensional vectors, $\epsilon \in \mathcal{N}(0,\sigma_y^2)$. Theorem~\ref{thm:BNN} considers the translation invariance, which motivates to ensure that using $\boldsymbol{w} + \eta\Delta\boldsymbol{w}$ (just like Eq.\eqref{eq:translation}) has the same likelihood with using $\boldsymbol{w}$. It is obvious that there exists a vector $\Delta\boldsymbol{w} \in \mathfrak{B}^{d-1}_1$, such that the following equations hold
	\[
	\left\{\begin{aligned}
		&\boldsymbol{w}^{\top} \boldsymbol{x} =
		\left( \boldsymbol{w} + \mathbf{B} \Delta\boldsymbol{w} \right)^{\top} \boldsymbol{x} \ , \\
		&\mathbf{B} = \begin{bmatrix}
			\mathbf{I}_{(d-1) \times (d-1)} \\
			- (x_1, \dots, x_{d-1}) / x_d
		\end{bmatrix} \ , \\
	\end{aligned}\right.
	\]
	which means that 
	\[
	\Delta\boldsymbol{w}^{\top} \mathbf{B}^{\top} \boldsymbol{x} =  
	(x_1, \dots, x_{d-1}, x_d)
	\begin{bmatrix}
		\Delta\boldsymbol{w} \\
		- \boldsymbol{x}^*_{d-1} \Delta\boldsymbol{w}
	\end{bmatrix}  = 0 \ ,
	\]
	where $\boldsymbol{x}^*_{d-1} = (x_1, \dots, x_{d-1}) / x_d$. Notice that $\eta$ is a singular value of the matrix $\mathbf{B}$. Provided a Gaussian prior $\boldsymbol{w} \in \mathcal{N}(\boldsymbol{\mu}_w,\boldsymbol{\Sigma}_w) $, we can approximate the likelihood from
	\begin{equation} \label{eq:likelihood}
		p(y \mid \boldsymbol{w}, \boldsymbol{x}) \approx q(\boldsymbol{w}) \propto \mathcal{N}(\boldsymbol{\mu}_x,\boldsymbol{\Sigma}_x) \ .    
	\end{equation}
	Thus, it suffices to show the marginal $q(\boldsymbol{w} + \mathbf{B} \Delta\boldsymbol{w})$ over $\Delta\boldsymbol{w} \in \mathcal{N}(\boldsymbol{0}, \beta^2\mathbf{I}_{(d-1) \times (d-1)}) $. For $\beta \to \infty$, we have
	\[
	\begin{aligned}
		\int q(\boldsymbol{w} + \mathbf{B} \Delta\boldsymbol{w}) p(\Delta\boldsymbol{w}) \dif (\Delta\boldsymbol{w}) 
		&= \int \mathcal{N}(\boldsymbol{\mu}_x,\boldsymbol{\Sigma}_x) \cdot \mathcal{N}(\boldsymbol{0}, \beta^2\mathbf{I}_{(d-1) \times (d-1)}) \dif (\Delta\boldsymbol{w})  \\
		&= \mathcal{N}(\boldsymbol{\mu}_x + \mathbf{B} ~\boldsymbol{0}_{(d-1)\ \times 1}, \boldsymbol{\Sigma}_x + \beta^2 \mathbf{B}\mathbf{B}^{\top}) \ ,
	\end{aligned}
	\]
	where the second equality holds from~\cite[Theorem 1]{kurle2022}. Thus, the inverse of the covariance matrix can be re-written as
	\[
	\begin{aligned}
		( \boldsymbol{\Sigma}_x + \beta^2 \mathbf{B}\mathbf{B}^{\top}  )^{-1} 
		&= \boldsymbol{\Sigma}_x^{-1} - \beta^{-2} \boldsymbol{\Sigma}_x^{-1} \mathbf{B} ( \mathbf{I}_{d-1} + \beta^2 \mathbf{B}^{\top} \boldsymbol{\Sigma}_x^{-1} \mathbf{B} )^{-1} \mathbf{B}^{\top} \boldsymbol{\Sigma}_x^{-1} \\
		&= \boldsymbol{\Sigma}_x^{-1} - \beta^{-2} \boldsymbol{\Sigma}_x^{-1} \mathbf{B} [ \beta^2 ( \beta^{-2} \mathbf{I}_{d-1} + \mathbf{B}^{\top} \boldsymbol{\Sigma}_x^{-1} \mathbf{B} ) ]^{-1} \mathbf{B}^{\top} \boldsymbol{\Sigma}_x^{-1} \\
		&= \boldsymbol{\Sigma}_x^{-1} - \boldsymbol{\Sigma}_x^{-1} \mathbf{B} ( \beta^{-2} \mathbf{I}_{d-1} + \mathbf{B}^{\top} \boldsymbol{\Sigma}_x^{-1} \mathbf{B} )^{-1} \mathbf{B}^{\top} \boldsymbol{\Sigma}_x^{-1} \ , 
	\end{aligned}
	\]
	where $\mathbf{I}_{(d-1)\times(d-1)}$ is abbreviated as $\mathbf{I}_{d-1}$.
	
	By building the likelihood model in Eq.~\eqref{eq:likelihood}, we can estimate the posterior as follows
	\begin{equation}  \label{eq:posterior}
		p(\boldsymbol{w} \mid \mathcal{D})
		\approx p(\boldsymbol{w} \mid \boldsymbol{x}) \propto q(\boldsymbol{w}) p(\boldsymbol{w}) = \mathcal{N}(\boldsymbol{\nu}, \boldsymbol{\Sigma}) \ ,    
	\end{equation}
	where
	\[
	\left\{\begin{aligned}
		\boldsymbol{\nu} & = \boldsymbol{\Sigma}_w(\boldsymbol{\Sigma}_x+\boldsymbol{\Sigma}_w)^{-1}\boldsymbol{\mu}_x + \boldsymbol{\Sigma}_x (\boldsymbol{\Sigma}_x+\boldsymbol{\Sigma}_w)^{-1} \boldsymbol{\mu}_w \ , \\
		\boldsymbol{\Sigma} &= \boldsymbol{\Sigma}_x (\boldsymbol{\Sigma}_x+\boldsymbol{\Sigma}_w)^{-1} \boldsymbol{\Sigma}_w \ ,
	\end{aligned}
	\right.
	\]
	which holds according to
	\[
	(\Sigma_w^{-1} + \Sigma_x^{-1})^{-1}
	= \Sigma_x(\Sigma_x + \Sigma_w)^{-1} \Sigma_w
	= \Sigma_w(\Sigma_w + \Sigma_x)^{-1} \Sigma_x \ .
	\]
	Combing Eq.~\eqref{eq:posterior} with Eq.~\eqref{eq:likelihood}, we have
	\[
	\begin{aligned}
		p(\boldsymbol{w} \mid \boldsymbol{x}) &\propto q(\boldsymbol{w}) p(\boldsymbol{w}) \\
		&\propto \mathcal{N}\left( \boldsymbol{\Sigma}_x^{-1} \boldsymbol{\mu}_x, \boldsymbol{\Sigma}_x^{-1} \right) \cdot \mathcal{N}\left( \boldsymbol{\Sigma}_w^{-1} \boldsymbol{\mu}_w, \boldsymbol{\Sigma}_w^{-1} \right) \\
		&= \mathcal{N}\left( \boldsymbol{\Sigma}_x^{-1} \boldsymbol{\mu}_x + \boldsymbol{\Sigma}_w^{-1} \boldsymbol{\mu}_w, \boldsymbol{\Sigma}_x^{-1} + \boldsymbol{\Sigma}_w^{-1} \right) \\
		&= \mathcal{N}(\boldsymbol{\nu}^*, \boldsymbol{\Sigma}^*) \ ,
	\end{aligned}
	\]
	where 
	\[
	\left\{\begin{aligned}
		\boldsymbol{\nu}^* & = ( \boldsymbol{\Sigma}_x^{-1} + \boldsymbol{\Sigma}_w^{-1} )^{-1} ( \boldsymbol{\Sigma}_x^{-1} \boldsymbol{\mu}_x + \boldsymbol{\Sigma}_w^{-1} \boldsymbol{\mu}_w  ) \ , \\
		\boldsymbol{\Sigma}^* &= ( \boldsymbol{\Sigma}_x^{-1} + \boldsymbol{\Sigma}_w^{-1} )^{-1}  \ .
	\end{aligned}
	\right.
	\]
	Since $\beta^2\mathbf{I}_{d-1}$ goes to zero as $\beta \to \infty$, thus one has
	\[
	p_{\infty}( \boldsymbol{w} \mid \boldsymbol{x}) = \mathcal{N}\left( \boldsymbol{\nu}^{**}, \boldsymbol{\Sigma}^{**}  \right) \ ,
	\]
	where
	\[
	\left\{\begin{aligned}
		\boldsymbol{\nu}^{**} &=  (\boldsymbol{\Sigma}_{\infty}^{-1} + \boldsymbol{\Sigma}_w^{-1})^{-1} ( \boldsymbol{\Sigma}_{\infty}^{-1} \boldsymbol{\mu}_x + \boldsymbol{\Sigma}_w^{-1} \boldsymbol{\mu}_w  ) \ , \\
		\boldsymbol{\Sigma}^{**} &= ( \boldsymbol{\Sigma}_{\infty}^{-1} + \boldsymbol{\Sigma}_w^{-1} )^{-1}  \ ,
	\end{aligned}\right.
	\]
	and
	\[
	\boldsymbol{\Sigma}_{\infty}^{-1} = \boldsymbol{\Sigma}^{-1} - \boldsymbol{\Sigma}^{-1} \mathbf{B} (\mathbf{B}^{\top} \boldsymbol{\Sigma}_x^{-1} \mathbf{B})^{-1} \mathbf{B}^{\top} \boldsymbol{\Sigma}^{-1} \ .
	\]
	
	The above over-parameterized linear probabilistic model can be easily extended to the concerned BNNs. Along this line, the problem that BNNs allowing the translation invariance coincides with solving the following equations that model all translation invariance of BNNs at some data point $\boldsymbol{x}$:
	\[
	\left\{\begin{aligned}
		\boldsymbol{z}^0 &= \boldsymbol{x} \in \mathcal{D} \ , \\
		\boldsymbol{z}^l &= \sigma( (\mathbf{W}^l + \mathbf{B}^l \Delta\mathbf{W}^l ) \boldsymbol{z}^{l-1} ),
		\quad\text{for}\quad l \in [L] \ ,
	\end{aligned}\right.
	\]
	where $d_l$ denotes the dimension of $\boldsymbol{z}^l$ and
	\[
	\mathbf{B}^l = \begin{bmatrix}
		\mathbf{I}_{(d_l-1) \times (d_l-1)} \\
		- (z^1, \dots, z^l_{d_l-1}) / z^l_{d_l}
	\end{bmatrix} \ .
	\]
	Similarly, the approximation likelihood can be calculated
	\begin{equation} \label{eq:itertive}
		\left\{\begin{aligned}
			q(\mathbf{W}^l) &= \mathbb{E}_{\boldsymbol{z}^{l-1}} \left[ \mathcal{N}\left( \boldsymbol{\nu}^*_l, \boldsymbol{\Sigma}^*_l  \right) \right]  \\
			p(\boldsymbol{z}^l) &= \mathbb{E}_{\boldsymbol{z}^{l-1}} \left[ \mathbb{E}_{\mathbf{W}^l \in q(\mathbf{W}^l)} \left[ \sigma( \mathbf{W}^l \boldsymbol{z}^{l-1} )  \right] \right]  \\
		\end{aligned}\right.    
	\end{equation}
	for $l \in [L]$, where
	\[
	\left\{\begin{aligned}
		\boldsymbol{\nu}^*_l &= \boldsymbol{u}_w + \frac{(\boldsymbol{z}^l)^{\top} (\boldsymbol{\mu}_x - \boldsymbol{\mu}_w) \boldsymbol{\Sigma}_w \boldsymbol{z}^l}{} \ , \\
		\boldsymbol{\Sigma}^*_l &=  \boldsymbol{\Sigma}_w - \frac{ \boldsymbol{\Sigma}~ (\boldsymbol{z}^l)^{\top} (\boldsymbol{z}^l)^{\top} \boldsymbol{\Sigma}^{\top} }{ \boldsymbol{\Sigma}^{\top} (\boldsymbol{\Sigma}_x + \boldsymbol{\Sigma}_w) \boldsymbol{z}^l }   \ . 
	\end{aligned}\right.
	\]
	It is observed that all the first and second moments (i.e., $\boldsymbol{\mu}_w, \boldsymbol{\mu}_x, \boldsymbol{\Sigma}_w, \boldsymbol{\Sigma}_x $ ) are constrained to the parts of the overall weight vector that corresponds to $\mathbf{W}^l$ and the expectation over $\boldsymbol{z}^l$ is taken w.r.t. $\boldsymbol{z}^l \in p(\boldsymbol{z}^{l-1})$ for $l \in [L]$.
	
	Therefore, we have proved that the function induced by BNNs is translation-invariant. Besides, the above invariant approximation can be optimized using iterative algorithms layer by layer, as shown in Appendix D. Thus, it is obvious that the optimization complexity of BNNs are less than that of DNNs, that is, $\mathcal{O} ( \alpha^{-T} )$ for $\alpha > 1$ and the iteration step indicator $T \in \mathbb{N}^+$. This completes the proof.
\end{proof}

\section{Applications}  \label{sec:applications}
The previous section showed the polynomial (asymptotic approximation and optimization) complexity of various neural networks to the invariant function and its variants. Inspired by this recognition, we here present a feasible application, which shows that exploiting the complexity advantages of neural networks can strengthen the effectiveness of the parameter estimation and forecasting of high-resolution signals. 

In general, we have some mild assumptions as follows:
\begin{assumption}
	We pre-process the signal data to ensure the following settings:
	\begin{itemize}
		\item[i)] There are $n$ sources $\{s_i\}_{i \in [n]}$ centered at frequency $\omega_0$;
		\item[ii)] Each source is a stationary zero-mean random process or deterministic signal;
		\item[iii)] Additive noise $\epsilon$ that belongs to a stationary zero-mean random process with covariance matrix $\Sigma_{\mathrm{noise}}$ is present at all $2m$ sensors, where $n \leq m$.
	\end{itemize}
\end{assumption}

Based on the above assumptions, the signals received at the $i$-th doublet can be formulated as
\begin{equation}  \label{eq:signal}
	\left\{\begin{aligned}
		p_i(t) &= \sum_{l=1}^n s_l(t) a_i(\theta_l) + \epsilon_{p,i}(t) \ , \\
		q_i(t) &= \sum_{l=1}^n s_l(t) \e^{j \omega_0 \sin(\theta_l) \delta / c} + \epsilon_{q,i}(t) \ ,
	\end{aligned}\right.   
\end{equation}
where $\theta_l$ is now the direction-of-arrival frequency of the $l$-th source relative to the direction of the translation displacement $\delta$. Correspondingly, Eq.~\eqref{eq:signal} has a vector-matrix formation
\[
\left\{\begin{aligned}
	\boldsymbol{p}(t) &= \mathbf{A} \boldsymbol{s}(t)  + \boldsymbol{\epsilon}_{p,:}(t) \ , \\
	\boldsymbol{q}(t) &= \mathbf{A} \mathbf{\Phi} \boldsymbol{s}(t) + \boldsymbol{\epsilon}_{q,:}(t) \ ,
\end{aligned}\right. 
\]
where $\boldsymbol{p} \in \mathbb{R}^m$, $\boldsymbol{s} \in \mathbb{R}^n$, $\mathbf{A} \in \mathbb{R}^{m \times n}$, and 
\[
\mathbf{\Phi} = \diag\{ \e^{j \omega_0 \sin(\theta_1) \delta / c}, \dots, \e^{j \omega_0 \sin(\theta_n) \delta / c} \}  \ .
\]
Provided the observed signals, we estimate the parameters $\mathbf{A}$ and $\mathbf{\Phi}$. Previously, there have been great efforts on this issue. The representative approaches contain the maximum likelihood (ML) method and maximum entropy (ME) method, which are often widely used but have certain limitations about the bias and sensitivity in parameter estimation~\cite{roy1986estimation}. Some researchers employed the statistical time series forecasting models, such as the auto-regressive moving average (ARMA) and auto-regressive integrated moving average (ARIMA) models~\cite{box2015}, to handle this issue. Besides, Schmidt and Bienvenu~\cite{barabell1998performance} independently exploited the measurement methods in the case of sensor arrays of arbitrary form, of which the more famous one is the multiple signal classification (MUSIC) model. Roy and Kailath~\cite{roy1989esprit} presented the ESPRIT by introducing rotational invariance to improve the procedure of convariance estimation, which reduces the computation and storage consumption in comparison with conventional studies including MUSIC.

Our method here is to improve the effectiveness of the parameter estimation and forecasting of high-resolution signals. The basic idea behind our method is to exploit the rotation invariance of the underlying signal subspace induced by the translation invariance of the sensor array, and then, to employ the FTNet — a special implementation beyond the complex-valued neural network, presented by Zhang and Zhou~\cite{zhang2021:FT} — to pursue the concerned signals.

To verify the effectiveness of our method, we conduct a simulation experiment with $n=10$ and $m=20$. Table~\ref{tab:paras} lists the comparative results between the classical estimation approaches and our method. It is observed that our method performs better with fewer parameters, which supports our proposed Theorem~\ref{thm:paras_CVNN} and recent advance~\cite[Theorem 3]{wu2021towards}.
\begin{table}
	\centering
	\caption{Parameter complexity and performance (Little is better) of comparative models for the task of forecasting high-resolution signals with $n=10$ and $m=20$.}
		\label{tab:paras}
		\begin{tabular}{@{}cccc@{}}
			\toprule
			Models & Settings & Paras & MSE \\ 
			\midrule
			MUSIC  & -- & -- & 25.37 \\
			ESPRIT & -- & -- & 23.14  \\
			VARIMA  & $(d,p,\Delta,q)=(20,10,1,5)$ & $7.0 \times 10^2$ & 25.70 \\
			FCN    & size(10,150,1)   & $1.6 \times 10^3$ & 21.09    \\
			RNN    & size(10,150,1)   & $2.4 \times 10^4$ & 19.32    \\
			LSTM   & size(10,150,1)   & $8.2 \times 10^4$ & 15.87    \\
			Our Work & size(10,150,1) & $3.0 \times 10^4$ & \textbf{13.33}  \\
			Our Work & size(10,50,1)  & $7.1 \times 10^3$ & 16.62  \\
			\bottomrule
	\end{tabular}
\end{table}

Finally, we also provide a discussion about the computational complexity. The primary computational advantage of ESPRIT is that it eliminates the search procedure inherent in all previous methods (ML, ME, MUSIC). ESPRIT produces signal parameter estimates directly in terms of (generalized) eigenvalues. As noted previously, this involves $\Theta(n^3)$ computations. On the other hand, other high-resolution techniques require a search over the parameter space, in which the exhaustive search is computationally expensive. The significant computational advantage of ESPRIT becomes even more pronounced in multidimensional parameter estimation where the computational load of ESPRIT grows linearly with dimension, while that of MUSIC grows exponentially. Let $r_i$ denote the number of vectors, the computation required to search over $m$ dimensions for $n$ parameter vectors is proportional to $ n \propto \prod_{i=1}^m r_i$. The results in Table~\ref{tab:paras} coincide with the above analyses.

\section{Conclusions, Discussions, and Prospects}  \label{sec:conclusions}
In this paper, we present the theoretical understandings of deep neural networks to invariant functions from the perspectives of approximation and complexity. We first show the universal approximation of neural networks to the generalized invariant functions, and then conclude that a broad range of invariant functions can be asymptotically approximated by various types of neural network models using a polynomial number of parameters or optimization iterations. We also provide a feasible application that connects the parameter estimation and forecasting of high-resolution signals with our theoretical results, which demonstrates the effectiveness of our method using simulation experiments.

Theorems~\ref{thm:paras_CVNN} and~\ref{thm:paras_CNN} show the advantages of the neural networks equipped with complex-valued or convolution operations in comparison with real-valued ones (from~\cite[Theorem 2]{zhang2021:cr}) via approximation complexity, where CVNNs and CNNs profit from unitary transformation and anti-symmetric multiplication, perhaps two most important characteristics of the complex-valued and convolutional neural networks. In Theorem~\ref{thm:paras_CVNN}, we adopt a considerably direct way, i.e., regard the relevance of the radius and phase to the target function as a prior (e.g., radial function) that imposes more constraints on a complex-valued neural network than a real-valued one, and thus, the unitary transformation that allows radius scaling and phase rotation (see Lemma~\ref{lemma:approximation-rotation}) yields an advantageous reduction of the approximation complexity. This means that merely doubling the number of real-valued parameters (or neurons) in each layer does not give the equivalent effect observed in a complex-valued neural network, which is consistent with the conclusions in~\cite{hirose2003,nils2018:doubling}. Alternatively, a complex number $z = z_1 + z_2 \ii$ can be written into a matrix form
\[
\begin{pmatrix}
	z_1 & -z_2 \\
	z_2 & z_1
\end{pmatrix} = z_1 \begin{pmatrix}
	1 & 0 \\
	0 & 1
\end{pmatrix} + z_2 \begin{pmatrix}
	0 & -1 \\
	1 & 0
\end{pmatrix} \ .
\]
Combined with the convolution operations, one has
\[
\begin{pmatrix}
	z_1 & -z_2 \\
	z_2 & z_1
\end{pmatrix} =  \begin{pmatrix}
	a & b \\
	c & d
\end{pmatrix} \cdot \begin{pmatrix}
	y_1 & -y_2 \\
	y_2 & y_1
\end{pmatrix} \ ,
\]
which yields an anti-symmetric multiplication~\cite{abraham1978} between the partial derivatives and independent variables, where $a,b,c,d \in \mathbb{R}$ and $y= y_1 + y_2 \ii$. Once half of entries are known, the other half are fixed~\cite{joshua2021:survy}. Such an adjoint relation also applies to reduce the complexity for approximating translation operations using convolutional neural networks (see Theorem~\ref{thm:paras_CNN}).

In light of the preceding merits, we conjecture that the neural networks equipped with complex-valued or convolution operations have the potential and power of representing the invariant functions with rotational and translation operations. Hence, our work may provide solid support for designating the theoretical legitimacy~\cite{trabelsi2019} and characterization~\cite{wu2021towards} of such network models. Besides, it would be interesting to theoretically study feature space transformation~\cite{zhou2021over} and width-depth representation~\cite{zhang2022:nngp} which might be keys to understanding mysteries behind the success of deep neural networks.


\appendix
This appendix provides the supplementary materials for our work ``On the Approximation and Complexity of Deep Neural Networks to Invariant Functions", constructed according to the corresponding sections therein. 

\section*{A. Full Proof of Lemma~\ref{lemma:closure}}
Let $\boldsymbol{x} \in K \subset \mathbb{R}^n$, $\tau \in \mathcal{G}$, and for arbitrary $\epsilon>0$ $g_\epsilon: \mathbb{R}^n \to \mathbb{R}$ is a $\mathcal{G}$-invariant function, satisfying that $\| f(\boldsymbol{x}) - g_\epsilon(\boldsymbol{x}) \|_{p,K} \leq \epsilon$. Then we have
\[
\begin{aligned}
	\| f(\boldsymbol{x}) - f(\tau(\boldsymbol{x})) \|_{p,K} 
	&\leq \| f(\boldsymbol{x}) - g_\epsilon(\boldsymbol{x}) \|_{p,K} + 
	\| g_\epsilon(\boldsymbol{x}) - g_\epsilon(\tau(\boldsymbol{x})) \|_{p,K} 
	+ \| g_\epsilon(\tau(\boldsymbol{x})) - f(\tau(\boldsymbol{x})) \|_{p,K} \\
	&\leq \epsilon + 0 + \epsilon = 2 \epsilon \ .
\end{aligned}
\]
This completes the proof. $\hfill\square$

\section*{B. Full Proof of Theorem~\ref{thm:paras_CVNN}} 
There are two proof methods for Theorem~\ref{thm:paras_CVNN}, one based on the rotation group action and one based on the Fourier transformation, which are detailed in this and the following sections, respectively.

\subsection*{B.1. Proof of Lemma~\ref{lemma:approximation-rotation}}
Let $f': \mathbb{R}^n \to \mathbb{R}$ be a radial function with $f'(\boldsymbol{x}) = \| \boldsymbol{x} \|$. According to the universal approximation theorems of CVNNs~\cite{voigtlaender2020}, such as~\cite[Theorem 1]{zhang2021:cr}, for any $\epsilon>0$, $i \in [n]$, and $d \geq 1$, one has
\begin{equation} \label{eq:12}	
	\sup_{\boldsymbol{x} \in \mathbb{C}^d} \Big| f'(\boldsymbol{x}) - [ \sigma(\boldsymbol{v_i}^{\top} \boldsymbol{x} + b_i) ]_{\mathrm{R}} \Big| \leq {\epsilon}/{2}  \ ,
\end{equation}
where $\boldsymbol{v_i} \in \mathbb{C}^{n}$ and $b_i \in \mathbb{C}$. Further, we define a new function $g': \mathbb{R} \to \mathbb{R}$ as follows
\[
g'(s) = \left[ \sum_{i=1}^{n'} \alpha_i' \sigma(s) + a' \right]_{\mathrm{R}} \ ,
\]
where $\alpha_i', a' \in \mathbb{R}$. For Lipschitz continuous function $\sqrt[r]{\cdot}$ and from~\cite[Lemma 1]{zhang2021:cr}, we have
\begin{equation} \label{eq:22_CVNN}
	\sup_{s\in[r^k,R^k]} \left| g(\sqrt[k]{s}) - g'(s) \right| \leq {\epsilon}/{2} \ ,
\end{equation}
where $n' \leq C' L(R^k-r^k) / (\sqrt[k]{r}\epsilon)$ for some constant $C'>0$ and integer $k \geq 2$. Further, we have
\begin{equation} \label{eq:32_CVNN}
	\left| g'(s) - f_{\textrm{CVNN}}(\boldsymbol{x}) \right| \leq \left| g'(s) - f'(\boldsymbol{x}) \right|
	+ \left| f'(\boldsymbol{x}) - f_{\textrm{CVNN}}(\boldsymbol{x}) \right| ,
\end{equation}
where $s = [ \boldsymbol{v_i}^{\top} \boldsymbol{x} + b_i ]_{\mathrm{R}}$ and 
\[
f'(\boldsymbol{x}) = \left[ \sum_{i=1}^{n'} w_i'~ \sigma(\boldsymbol{v_i}^{\top} \boldsymbol{x} + b_i) + a  \right]_{\mathrm{R}}  \ ,
\]
in which $\{w_i', a\}$ denotes another collection of weighted parameters that corresponds to 
\[
f_{\textrm{CVNN}}(\boldsymbol{x}) = \left[ \sum_{i\in[n]} w_i \sigma(\boldsymbol{v_i}^{\top} \boldsymbol{x} + b_i) + a \right]_{\mathrm{R}} \ ,
\]
The first term of Eq.~\eqref{eq:32_CVNN} can be bounded by $\epsilon /4$ from~\cite[Lemma 1]{zhang2021:cr} for any $s \in [r^k, R^k]$. The second term is at most $\epsilon /4$ when $m \geq m'$ from Eq.~\eqref{eq:12}. This follows that
\begin{equation} \label{eq:42_CVNN}
	\left| g'(s) - f_{\textrm{CVNN}}(\boldsymbol{x}) \right| \leq  \epsilon / 2 \ .
\end{equation}
Combining with Eqs.~\eqref{eq:22_CVNN} and~\eqref{eq:42_CVNN}, we have
\[
\left| g(\|\boldsymbol{x}\|) - f_{\textrm{CVNN}}(\boldsymbol{x}) \right|
\leq \left| g(\sqrt[k]{s}) - g'(s) \right|   +  \left| g'(s) - f_{\textrm{CVNN}}(\boldsymbol{x}) \right| 
\leq \epsilon  \ ,
\]
where $\boldsymbol{x}\in\mathbb{C}^{d}$ and $s \in [r^k, R^k]$. We finally obtain
\[
m \leq C_s (R^k-r^k) Ld / (\sqrt[k]{r}\epsilon) \ ,
\]
provided $m \leq dm'$ and $C' \leq C_s$. Finally, we can complete the proof by setting $k=2$ in the above upper bound. $\hfill\square$

\subsection*{B.2. Proof of Lemma~\ref{lemma:L-rotation}}
Let $r = C_2\sqrt{d}$, $R = 2C_2\sqrt{d}$, and $d \geq 1$, then we have $r\geq 1$, which satisfies the condition of~\cite[Lemma 1]{zhang2021:cr}. Invoke Lemma~\ref{lemma:approximation-rotation} to construct the concerned CVNN and define $\epsilon' \leq \epsilon / d$. Then for any $L$-Lipschitz radial function $g:\mathbb{C}^{d} \to \mathbb{R}$ supported on $\mathcal{S}_{\Delta}$, we have
\[
\sup_{\boldsymbol{x}\in\mathbb{C}^{d}} | g(\boldsymbol{x}) - f_{\textrm{CVNN}}(\boldsymbol{x}) | \leq \epsilon' \ ,
\]
where the width of the hidden layer is bounded by
\[
m \leq \frac{C_s(C_2)^{3/2}dL}{\epsilon}d^{3/4} \leq \frac{C_s(C_2)^{3/2}L}{\epsilon}d^{7/4} \ .
\]
This completes the proof. $\hfill\square$

\subsection*{A.3. Proof of Lemma~\ref{lemma:define_g}}
Define a branch function
\[
h_i(\boldsymbol{x}) = \left\{ \begin{aligned}
	\max\{ \mathbb{I}\{\|\boldsymbol{x}\|\in \Omega_i\}, ND_i \}, &\quad\text{if}\quad B_i = 1, \\
	0\quad\quad\quad, &\quad\text{if}\quad B_i = 0,
\end{aligned}\right.
\]
where $B_i$ denotes a binary indicator and
\[
D_i = \min\left\{
\left|~ \|\boldsymbol{x}\| - \left( 1+ \frac{i-1}{N} \right) C_2 \sqrt{d} ~\right|, 
\left|~ \|\boldsymbol{x}\| - \left( 1+ \frac{i}{N} \right) C_2 \sqrt{d}  ~\right| \quad\right\} .
\]
Let $h(\boldsymbol{x})  = \sum_{i=1}^N \beta_i h_i(\boldsymbol{x})$, where $\beta_i \in \{-1, + 1\}$, $B_i = 1$, $\Omega_i$'s are disjoint intervals, $h_i(\boldsymbol{x})$ is an $N$-Lipschitz function. Thus, $h$ is also an $N$-Lipschitz function. So we have
\[
\begin{aligned}
	\int_{\mathbb{C}^{d}} \left( h(\boldsymbol{x}) - \sum_{i=1}^N \epsilon_i g_i(\boldsymbol{x}) \right)^2 \phi^2(\boldsymbol{x}) \dif\boldsymbol{x} 
	& = \int_{\mathbb{C}^{d}}  \sum_{i=1}^N \epsilon_i^2 \left( h_i(\boldsymbol{x}) - g_i(\boldsymbol{x}) \right)^2 \phi^2(\boldsymbol{x}) \dif\boldsymbol{x} \\
	& = \sum_{i=1}^N \int_{\mathbb{C}^{d}} \left( h_i(\boldsymbol{x}) - g_i(\boldsymbol{x}) \right)^2 \phi^2(\boldsymbol{x}) \dif\boldsymbol{x} \\
	& \leq \frac{3}{(C_2)^2\sqrt{d}} \ ,
\end{aligned}
\]
where the last inequality holds from \cite[Lemma 22]{eldan2016}. This completes the proof. $\hfill\square$

Finally, we can complete the proof of Theorem~\ref{thm:paras_CVNN}.

\vspace{0.1 cm}
\noindent\textit{Finishing the Proof of Theorem~\ref{thm:paras_CVNN}.}\quad Let $g(\boldsymbol{x}) = \sum_{i=1}^N \epsilon_i g_i(\boldsymbol{x})$ be defined by Eq.~\eqref{eq:target_radial} and $N \geq 4C_2^{5/2}d^2$. According to Lemma~\ref{lemma:define_g}, there exists a Lipschitz function $h$ with range $[-1,+1]$ such that
\[
\left\| h(\boldsymbol{x}) - g(\boldsymbol{x}) \right\|_{L_2(\mu)} \leq \frac{\sqrt{3}}{C_2 d^{1/4}} \ .
\]
Based on Lemmas~\ref{lemma:approximation-rotation} and~\ref{lemma:L-rotation}, any Lipschitz radial function supported on $\mathcal{S}_{\Delta}$ can be approximated by an expressive function $f_{\textrm{CVNN}}(\boldsymbol{x})$ led by a one-hidden-layer CVNN with width at most $C_3C_sd^{15/4} / \epsilon$, where $C_3$ is a constant relative to $C_2$ and $\epsilon$. This means that
\[
\sup_{\boldsymbol{x}\in\mathbb{C}^{d}} | h(\boldsymbol{x}) - f_{\textrm{CVNN}}(\boldsymbol{x}) | \leq \epsilon \ .
\]
Thus, we have
\[
\| h(\boldsymbol{x}) - f_{\textrm{CVNN}}(\boldsymbol{x}) \|_{L_2(\mu)} \leq \epsilon \ .
\]
Hence, the range of $f_{\textrm{CVNN}}(\boldsymbol{x})$ is in $[-1-\epsilon, +1+\epsilon] \subseteq [-2, +2]$ given $\epsilon \leq 1$. For the radial function in Eq.~\eqref{eq:target_radial}, we have
\[
\| g(\boldsymbol{x}) - f_{\textrm{CVNN}}(\boldsymbol{x}) \|_{L_2(\mu)}
	 \leq \| g(\boldsymbol{x}) - h(\boldsymbol{x}) \|_{L_2(\mu)}  + \| h(\boldsymbol{x}) - f_{\textrm{CVNN}}(\boldsymbol{x}) \|_{L_2(\mu)} 
	 \leq \frac{\sqrt{3}}{C_2 d^{1/4}} + \epsilon \ .
\]
This implies that given constants $m > C_2 >0$ and $C_3>0$, for any $\epsilon>0$ and $\beta_i \in \{-1,+1\}$ $(i\in[N])$, the target radial function $g$ can be approximated by an expressive function $f_{\textrm{CVNN}}(\boldsymbol{x})$ led by a one-hidden-layer CVNN with range in $[-2,+2]$ and width at most $C_3C_sd^{15/4} / \epsilon$, that is,
\[
\| g(\boldsymbol{x}) - f_{\textrm{CVNN}}(\boldsymbol{x}) \|_{L_2(\mu)} \leq \frac{\sqrt{3}}{C_2 d^{1/4}} + \epsilon  \ .
\]
This completes the proof.  $\hfill\square$

\section*{C. Another Proof for Theorem~\ref{thm:paras_CVNN}}
The proof idea can be summarized as follows. Given any permutation operation on the input channels, there are some invariant properties on the complex-valued operations. In other words, the expressive function led by a one-hidden-layer CVNN admits the rotation transformations. Therefore, it suffices to show that before final weighted aggregation, the component expressive functions of hidden neurons can approximate any $L$-Lipschitz radial function. Then we can find a special radial function that can be well approximated by these hidden expressive functions within the polynomial parameter complexity.

\begin{lemma} \label{lemma:x_rotation}
	Let $\sigma$ and $f_{\mathrm{CVNN}}(\boldsymbol{x})$ be an $l$-finite function and the expressive function led by a one-hidden-layer CVNN, respectively. For any $\epsilon>0$ and $\mathbf{A} \in \mathbf{SO}(n,\mathbb{R})$, there exists some time such that
	\[
	\left| f_{\mathrm{CVNN}}(\mathbf{A}\boldsymbol{x}) - \| \boldsymbol{x} \| \right| \leq \epsilon  \ .
	\]
\end{lemma}

\begin{lemma} \label{lemma:g_rotation}
	Let $\sigma$ be an $l$-finite function, $g: \mathbb{R} \to \mathbb{R}$ is a scalar function, and $f_{\mathrm{CVNN}}(\boldsymbol{x})$ is the expressive function led by a one-hidden-layer CVNN, where $\boldsymbol{x} \in \mathcal{S}(r)$ and $\mathcal{S}(r)$ is a sphere supported with density $\phi^2$ for $0<r<\infty$. For any $\epsilon>0$, it holds
	\[
	|f_{\mathrm{CVNN}}(\boldsymbol{x}) - g(\|\boldsymbol{x}\|)| \leq \epsilon
	\]
	if and only if
	\[
	\left| f_{\mathrm{CVNN}}(\mathbf{A}\boldsymbol{x}) - g(\| \boldsymbol{x} \|) \right| \leq \epsilon  \ ,
	\]
	for any $\mathbf{A} \in \mathbf{SO}(n,\mathbb{R})$.
\end{lemma}

\begin{lemma} \label{lemma:app_complex_CVNN}
	Define a radial function $g': \mathbb{R} \to \mathbb{R}$ with the form of
	\[
	g'(\|\boldsymbol{x}\|) \triangleq \sum_{i=1}^N \beta_i ~\mathbb{I}\{\|\boldsymbol{x}\|\in \Omega_i\} \ ,
	\]
	where $N$ is a polynomial function of $d$, $\boldsymbol{\beta} = (\beta_1,\dots,\beta_N) \in \{-1, +1\}^N$, and $\Omega_i$'s are disjoint intervals of width $\mathcal{O}(1/N)$ on values in the range $\Theta(\sqrt{d})$. For $C_2, C_3>0$ with $d>C_2$, any $\epsilon>0$, and any choice of $\epsilon_i \in \{-1,+1\}$ $(i\in[N])$, there exists some matrix $\mathbf{A} \in \mathbf{SO}(n,\mathbb{R})$ such that
	\[
	\left| g'(\|\boldsymbol{x}\|) - f_{\mathrm{CVNN}}(\mathbf{A}\boldsymbol{x}) \right| \leq \frac{\sqrt{3}}{C_2 d^{1/4}} + \epsilon \ ,
	\]
	where $f_{\mathrm{CVNN}}(\boldsymbol{x})$ indicates the expressive function led by a one-hidden-layer CVNN with range in $[-2,+2]$ and at most $C_3C_sd^{15/4} / \epsilon$ hidden neurons.
\end{lemma}

\noindent\textit{Finishing the Proof of Theorem~\ref{thm:paras_CVNN}}\quad The rest proof is a straightforward combination of Lemmas~\ref{lemma:x_rotation}, \ref{lemma:g_rotation} and~\ref{lemma:app_complex_CVNN}. Define a radial function $g': \mathbb{R} \to \mathbb{R}$ with the form of
\[
g'(\|\boldsymbol{x}\|) \triangleq \sum_{i=1}^N \beta_i ~\mathbb{I}\{\|\boldsymbol{x}\|\in \Omega_i\} \ ,
\]
where $N$ is a polynomial function of $m$, $\boldsymbol{\beta} = (\beta_1,\dots,\beta_N) \in \{-1, +1\}^N$, and $\Omega_i$'s are disjoint intervals of width $\mathcal{O}(1/N)$ on values in the range $\Theta(\sqrt{d})$. Obviously, $g'$ is equivalent to another radial function defined in Eq.~\eqref{eq:target_radial} as follows
\[
g(\boldsymbol{x}) = \sum_{i=1}^N \beta_i g_i(\boldsymbol{x})
\quad \text{ with }\quad
g_i(\boldsymbol{x}) = \mathbb{I}\{\|\boldsymbol{x}\|\in \Omega_i\} \ ,
\]
provided $\boldsymbol{x} \in \mathcal{S}(r)$. From Lemma~\ref{lemma:app_complex_CVNN}, it is observed that provided the concerned radial function $g(\boldsymbol{x})$ with a collection of choices $\boldsymbol{\beta}$, there exists $m \in \mathcal{O}(C_3C_sd^{15/4} / \epsilon)$ such that $g(\boldsymbol{x})$ can be approximated by an expressive function $f_{\textrm{CVNN}}(\boldsymbol{x})$ led by a one-hidden-layer CVNN with range in $[-2,+2]$ and at most $m$ hidden neurons, such that
\[
\left| g'(\|\boldsymbol{x}\|) - f_{\textrm{CVNN}}(\mathbf{A}\boldsymbol{x}) \right| \leq \frac{\sqrt{3}}{C_2 d^{1/4}} + \epsilon \ .
\]
Let $\mathbf{A} \in \mathbf{SO}(n,\mathbb{R})$. If $\|\boldsymbol{x}\| \in \Omega_i \subseteq \mathcal{S}(r)$ for $0 < r < \infty$, then it holds $\|\mathbf{A} \boldsymbol{x}\| \in \Omega_i \subseteq \mathcal{S}(r)$ and $g'( \|\boldsymbol{x}\|) = g'(\|\mathbf{A}\boldsymbol{x}\|)$. According to Lemma~\ref{lemma:g_rotation}, we have 
\[
\begin{aligned}
	\left\| g(\boldsymbol{x}) - f_{\textrm{CVNN}}(\boldsymbol{x},t) \right\|_{L_2(\mu)}
	&= \left| g'( \|\boldsymbol{x}\|) - f_{\textrm{CVNN}}(\boldsymbol{x}) \right| \\
	&= \left| g'(\|\boldsymbol{x}\|) - f_{\textrm{CVNN}}(\mathbf{A}\boldsymbol{x}) \right| \\
	&\leq \frac{\sqrt{3}}{C_2 d^{1/4}} + \epsilon \ ,
\end{aligned}
\]
without any incremental change in the parameter complexity. This completes the proof.  $\hfill\square$

\section*{D. Iterative Algorithms for Bayesian Neural Networks} \label{app:BNN} 
Here, we show the iterative algorithm for optimizing the invariant approximation using BNNs layer by layer in Subsection~\ref{subsec:BNN}. The procedure is listed as follows:
\begin{itemize}
	\item Provided data $\boldsymbol{x}$, using Eq.~\eqref{eq:itertive} and the point estimation for the weights in the $l$-th layer to optimize the parameters (i.e., mean vector and covariance matrix) of the weights in the $(l-1)$-th layer for $2 \leq l \leq L$;
	\item One generates distribution $p$ of activation variables $\boldsymbol{z}^l$ composed of the $l$-th layer by exploiting Eq.~\eqref{eq:itertive} under the optimal $q(\mathbf{W}^l)$;
	\item The above steps can be repeated until the last layer, (that is, $l = L$), and then $q(\mathbf{W}^l)$ converges over iteration. 
\end{itemize}
It is observed that this iterative procedure is no larger than optimizing the weights of DNNs, since all previous layers are full-characterized by the distribution of the latent activation variables $\boldsymbol{z}^l$ of the hidden layers.

\bibliography{JMref}
\bibliographystyle{plain}

\end{document}